# Quantitative Evaluation of Performance and Validity Indices for Clustering the Web Navigational Sessions

Zahid Ansari[1], M.F. Azeem[3], Waseem Ahmed[4]
[1,4]Dept. of Computer Science, [3]Dept. of Electronics
P.A. College of Engineering
Mangalore, India

A.Vinaya Babu[2]
Dept. of Computer Science Engineering
Jawaharlal Nehru Technical University
Hyderabad, India

*Abstract*—Clustering techniques are widely used in "Web Usage Mining" to capture similar interests and trends among users accessing a Web site. For this purpose, web access logs generated at a particular web site are preprocessed to discover the user navigational sessions. Clustering techniques are then applied to group the user session data into user session clusters, where inter-cluster similarities are minimized while the intra cluster similarities are maximized. Since the application of different clustering algorithms generally results in different sets of cluster formation, it is important to evaluate the performance of these methods in terms of accuracy and validity of the clusters, and also the time required to generate them, using appropriate performance measures. This paper describes various validity and accuracy measures including Dunn's Index, Davies Bouldin Index, C Index, Rand Index, Jaccard Index, Silhouette Index, Fowlkes Mallows and Sum of the Squared Error (SSE). We conducted the performance evaluation of the following clustering techniques: *k*-Means, *k*-Medoids, Leader, Single Link Agglomerative Hierarchical and DBSCAN. These techniques are implemented and tested against the Web user navigational data. Finally their performance results are presented and compared.

## I. INTRODUCTION

Clustering techniques are widely used in WUM to capture similar interests and trends among users accessing a Web site. Clustering aims to divide a data set into groups or clusters where inter-cluster similarities are minimized while the intra cluster similarities are maximized. Details of various clustering techniques can be found in survey articles [1]-[3]. Some of the main categories of the clustering methods are [4]: i) *Partitioning* methods, that create k partitions of a given data set, each representing a cluster. Typical partitioning methods include k-means, k-medoids etc. In *k-means* algorithm each cluster is represented by the mean value of the data points in the cluster called centroid of the cluster. On the other hand and in *k-medoids* algorithm, each cluster is represented by one of the data point located near the center of the cluster called medoid of the cluster. Leader clustering is also a partitioning based clustering techniques which generates the clusters based on an initially specified dissimilarity measure, ii) *Hierarchical* methods create a hierarchical decomposition of the given set of data objects. A hierarchical method can be classified as being either agglomerative or divisive, based on how the hierarchical decomposition is formed. iii) *Density- based* methods form the clusters based on the notion of density. They can discover the clusters of arbitrary shapes. These methods continue growing the given cluster as long as the number of objects or data points in the "neighborhood" exceeds some threshold. DBSCAN is a typical density-based method that grows clusters according to a density-based connectivity analysis.

A number of clustering algorithms have been used in Web usage mining where the data items are user sessions consisting of sequence of page URLs accessed and interest scores on each URL page based on the characteristics of user behavior such as time elapsed on a page or the bytes downloaded [5]. In this context, clustering can be used in two ways, either to cluster users or to cluster items. In user-based clustering, users are grouped together based on the similarity of their web page navigational patterns. In item based clustering, items are clustered based on the similarity of the interest scores for these items across all users [6], [7]. Since the application of different clustering algorithms generally results in a different set of cluster formation, it is important to perform an evaluation of the performance of these methods in terms of accuracy and validity of the clusters, and also the time required to generate them, using appropriate performance measures.

The remainder of the paper is organized as follows. Section II presents a overview of various clustering techniques used for mining web user sessions clusters, including k-Means, k-Medoids, Leader, Hierarchical and DBSCAn. Section III presents various validity indices to judge the validity of the clusters formed by various techniques. Section IV describes the experimental results and finaly conclusion is provided in Section V.

## II. Techniques Used for Clustering User Sessions

Each user session is mapped as a vector of URL references in an $n$-dimensional space. Let $U = \{u_1, u_2, \ldots, u_n\}$ be a set of $n$ unique URLs appearing in the preprocessed log and $S = \{s_1, s_2, \ldots, s_m\}$ be a set of $m$ user sessions discovered by preprocessing the web log data, where each user session $s_i \in S$ can be represented as $s = \{w_{u_1}, w_{u_2}, \ldots, w_{u_m}\}$. Here $w_{u_i}$ may take either a binary or non-binary value depending on whether it represents presence and absence of the URL in the session or some other feature of the URL. For this work, $w_{u_i}$ represents the time a user spends on the URL item $u_i$.

Remainder of this section provides a detailed discussion of each clustering technique used for discovering the user session clusters.

### A. k-Means Clustering Algorithm:

The *k*-Means clustering algorithm [8] is one of the most commonly used methods for partitioning the data. Given a set of $m$ data points $X = \{x_i \mid i = 1 \cdots m\}$, where each data point is a $n$-dimensional vector, *k*-means clustering algorithm aims to partition the $m$ data points into $k$ clusters $C = \{c_1, c_2, \ldots, c_k\}$, so as to minimize an objective function $J(V, X)$ of dissimilarity [9], which is the within-cluster sum of squares. The objective function $J$, based on the Euclidean distance between a data point vector $x_i$ in cluster $j$ and the corresponding cluster center $v_j$, is defined in (1).

$$J(X,V) = \sum_{j=1}^{k} J_i(x_i, v_j) = \sum_{j=1}^{k} \left( \sum_{i=1}^{m} u_{ij} \cdot d^2(x_i, v_j) \right), \quad (1)$$

where, $J_i(x_i, v_j) = \sum_{i=1}^{m} u_{ij} \cdot d^2(x_i, v_j)$,

is the objective function within cluster $c_i$,

$u_{ij} = 1$, if $x_i \in c_j$ and 0 otherwise.

$d^2(x_i, v_j)$ is the distance between $x_i$ and $v_j$

$$d^2(x_i, v_j) = \left\| \sum_{k=1}^{n} x_k^i - v_k^j \right\|^2$$

where, $n$ is the number of dimensions of each data point

$x_k^i$ is the value of $k^{th}$ dimensions of $x_i$

$v_k^j$ is the value of $k^{th}$ dimensions of $v_j$

The partitioned clusters are defined by a $m \times k$ binary membership matrix $U$, where the element $u_{ij}$ is 1, if the data point $x_i$ belongs to the cluster $j$, and 0 otherwise. Once the cluster centers $V = \{v_1, v_2, \ldots, v_k\}$, are fixed, the membership function $u_{ij}$ that minimizes (1) can be derived as follows:

$$u_{ij} = \begin{cases} 1; & \text{if } d^2(x_i, v_j) \leq d^2(x_i, v_{j*}) \ j \neq j^*, \forall \ j^* = 1, \cdots, k \\ 0; & \text{otherwise} \end{cases} \quad (2)$$

The equation (4) specifies that assign each data point $x_i$ to the cluster $c_j$ with the closest cluster center $v_j$. Once the membership matrix $U = [u_{ij}]$ is fixed, the optimal center $v_j$ that minimizes (1) is the mean of all the data point vectors in cluster $j$ can be computed using (3).

$$v_j = \frac{1}{|c_j|} \sum_{i, x_i \in c_j}^{m} x_i \quad (3)$$

where,

$|c_j|$, is the size of cluster $c_j$ and also $|c_j| = \sum_{i=1}^{m} u_{ij}$

Given an initial set of $k$ means or cluster centers, $V = \{v_1, v_2, \ldots, v_k\}$, the algorithm proceeds by alternating between two steps: i) Assignment step: Assign each data point to the cluster with the closest cluster center. ii) Update step: Update the cluster center as the mean of all the data points in that cluster. The input to the algorithm is a set of $m$ data points $X = \{x_i \mid i = 1 \cdots m\}$, where each data point is a $n$-dimensional vector, it then determines the cluster centers $v_j$ and the membership matrix $U$ iteratively as explained in Fig. 1.

---

**Algorithm:** *k*-Means clustering algorithm for partitioning, where each cluster's center is represented by the mean value of the data points in that cluster.

**Input:** $k$, the number of clusters and Set of $m$ data points $X = \{x_1, \ldots, x_m\}$.

**Output:** Set of $k$ centroids, $V = \{v_1, \ldots, v_k\}$, corresponding to the clusters $C = \{c_1, \ldots, c_k\}$, and membership matrix $U = [u_{ij}]$.

**Steps:**

1) Initialize the $k$ centroids $V = \{v_1, \ldots, v_k\}$, by randomly selecting $k$ data points from $X$.

2) **repeat**

   i) Determine the membership matrix $U$ using (2), by assigning each data point $x_i$ to the closest cluster $c_j$.

   ii) Compute the objective function $J(X,V)$ using (1). Stop if it below a certain threshold $\varepsilon$.

   iii) Recompute the centroid of each cluster using (3).

3) **until** Centroids do not change

---

Figure 1. *k*-Means Clustering Algorithm

### B. K-Medoids Clustering Algorthm:

*k*-Medoid is a classical partitioning technique of clustering that clusters the data set of $m$ data points into $k$ clusters. It attempts to minimize the squared error, which is the distance between data points within a cluster and a point designated as the center of that cluster. In contrast to the *k*-means algorithm, *k*-Medoids algorithm selects data points as cluster centers (or medoids). A medoid is a data point of a cluster, whose average dissimilarity to all the other data points in the cluster is minimal i.e. it is a most centrally located data point in the cluster [10],[11].

Given a set of $m$ data points $X = \{x_i \mid i = 1 \cdots m\}$, where each data point is a $n$-dimensional vector, $k$-mdoids clustering algorithm partitions the $m$ data points into $k$ clusters, so as to minimize an objective function representing the sum of the dissimilarities between each of the data points and its corresponding cluster medoid. Let $M = \{m_1, m_2, \ldots, m_k\}$ be the set of medoids corresponding to $C$. The objective function $J(X, M)$ is defined in (4)

$$J(X,M) = \sum_{j=1}^{k} \left( \sum_{i=1}^{m} u_{ij} \cdot d^2(x_i, m_j) \right), \quad (4)$$

where,

$x_i$ is the $i^{th}$ data point

$m_j$ is the medoid of cluster $c_j$

$u_{ij} = 1$, if $x_i \in c_j$ and 0 otherwise.

$d^2(x_i, m_j)$ is the Euclidean distance between $x_i$ and $m_j$

$$d^2(x_i, m_j) = \left\| \sum_{k=1}^{n} x_k^i - m_k^j \right\|^2$$

where, $n$ is the number of dimensions of each data point

The membership function $u_{ij}$ that minimizes (4) can be derived as follows:

$$u_{ij} = \begin{cases} 1; & \text{if } d^2(x_i, m_j) \le d^2(x_i, m_{j*}) \; j \ne j^*, \forall \; j^* = 1, \cdots, k \\ 0; & \text{otherwise} \end{cases} \quad (5)$$

Once the membership matrix $U = [u_{ij}]$ is fixed, the new cluster medoids $m_j$ that minimizes (4) can be found using (6).

$$m_j = \arg\min_{x_i \in c_j} \sum_{x_l \in c_j} d(x_i, x_l) \quad (6)$$

The most common realisation of $k$-medoid clustering is the Partitioning Around Medoids (PAM) algorithm and is as described in Fig 2.

---

**Algorithm:** *k*-Medoids Clustering

**Input:** Set of $m$ data points $X = \{x_1, \ldots, x_m\}$.

**Output:** Set of $k$ medoids, $M = \{m_1, \ldots, m_k\}$, corresponding to the clusters $C = \{c_1, \ldots, c_k\}$, and membership matrix $U = [u_{ij}]$ that minimizes the sum of the dissimilarities of all the data points to their nearest medoid.

**Steps:**

4) Initialize the $k$ medoids $V = \{v_1, \ldots, v_k\}$, by randomly selecting $k$ data points from $X$.

5) **repeat**

   iv) Determine the membership matrix $U$ using (5), by assigning each data point $x_i$ to the closest cluster $c_j$.

   v) Compute the objective function $J(X,M)$ using (4). Stop if it below a certain threshold $\varepsilon$.

   vi) Recompute the medoid of each cluster using (6).

6) **until** Medoids do not change

---

Figure 2. *k*-Medoids Clustering Algorithm

### C. Leader Clustering Algorithm:

The leader clustering algorithm [12],[13] is based on a predefined dissimilarity threshold. Initially, a random data point from the input data set is selected as leader. Subsequently, distance of every other data point with the selected leader is computed. If the distance of a data point is less than the dissimilarity threshold, that data point falls in the cluster with the initial leader. Otherwise, the data point is identified as a new leader. The computation of leaders is continued till all the data points are considered. The number of leaders is inversely proportional to the selected threshold.

Given a set of $m$ data points $X = \{x_i \mid i = 1 \cdots m\}$, where each data point is a $n$-dimensional vector. The Euclidean distance between the $i^{th}$ data point $x_i \in X$ and $j^{th}$ leader $l_j \in L$ (where $L$ is a set of leaders) is given by:

$$d^2(x_i, l_j) = \left\| \sum_{k=1}^{n} x_k^i - l_k^j \right\|^2 \quad (7)$$

where, $n$ is the number of dimensions of each data point

$x_k^i$ is the value of $k^{th}$ dimensions of $x_i$

$l_k^j$ is the value of $k^{th}$ dimensions of $x_j$

The leader clustering algorithm is described in Fig. 3.

---

**Algorithm:** Leader Clustering

**Input:**   i) Set of $m$ data points $X = \{x_1, \ldots, x_m\}$,
          ii) α, the dissimilarity threshold.

**Output:** Set of clusters $C = \{c_1, \ldots, c_k\}$,

**Steps:**

1) $C = \phi, L = \phi, j = 1$    // Initilize the cluster and leader sets

2) $l_j = x_1$             // Initialize $x_1$ as the first leader

3) $L = L \cup l_j$

4) $c_j = c_j \cup x_1$

5) $C = C \cup c_j$

6) **for each** $x_i \in X$ where $i = 2, \ldots m$

7) **begin**

8)     $j = \arg\min_{j, \; l_j \in L} d(x_i, l_j)$

9)     **if** $d^2(x_i, l_j) < \alpha$ **then**

10)    $c_j = c_j \cup x_i$

11)    **else**

12)     $j = j + 1$

13)     $l_j = x_i$

14)     $L = L \cup l_j$

15)     $c_j = c_j \cup x_i$

16)     $C = C \cup c_j$

17)    **endif**

18) **end**

---

Figure 3. Leader Clustering Algorithm

*D. Agglomerative Hierarchichal Clustering*

Agglomerative Hierarchical clustering method groups the data objects into a tree of clusters. The hierarchical clusters are formed in a bottom-up fashion. It starts by placing each individual data point in its own cluster and then merges these atomic clusters into larger and larger clusters. This process continues until all of the data points gather in a single cluster or certain termination conditions are satisfied [4]. Clusters are agglomerated based on the similarity measure between the two clusters. Some of the most widely used measures for computing the distance between a pair of clusters are given below:

$$d_{min}(c_i, c_j) = \min_{x_1 \in c_i, x_2 \in c_j} d(x_1, x_2) \quad (8)$$

$$d_{max}(c_i, c_j) = \max_{x_1 \in c_i, x_2 \in c_j} d(x_1, x_2) \quad (9)$$

$$d_{avg}(c_i, c_j) = \frac{1}{m_i m_j} \sum_{x_1 \in c_i} \sum_{x_2 \in c_j} d(x_1, x_2) \quad (10)$$

where $c_i$ and $c_j$ represent clusters $i$ and $j$ containing $m_i$ and $m_j$ number of data points respectively.

The two most popular hierarchical clustering algorithms are single-link [14], complete-link [15]. These algorithms differ in the way they measure the distance between a pair of clusters [16]. In the single link method, the distance measure between two clusters is the minimum of the distances between all pairs of data points from the two clusters as defined in (8). In the complete link algorithm, the distance between two clusters is the maximum of all pair wise distances between data points in the two clusters as described in (9). The basic single link algorithm is given below:

---

**Algorithm:** Single Link Agglomerative Hierarchical Clustering

**Input:** Set of $m$ data points $X=\{x_1, …, x_m\}$.

**Output:** A single cluster $\{c\}$
**Steps:**

1) Compute the $m \times m$ proximity matrix $D$ containing distances $d(x_i, x_j) \forall i, j = 1, …, m$

2) **repeat**

   i) Find the most similar pair of clusters $c_i$ and $c_j$ using (8)

   ii) Merge clusters $c_i$ and $c_j$ into a single cluster.

   iii) Update the proximity matrix $D$, by deleting the rows and columns corresponding to $c_i$ and $c_j$ and adding a row and column corresponding to the newly formed cluster. The proximity between this new cluster c, and an old cluster k is defined as:

   $$d_{min}(c, c_k) = \min d_{min}(c_i, c_k), d_{min}(c_j, c_k)$$

3) **until** only one cluster remains

---

Figure 4. Single Link Agglomerative Clustering Algorithm

*E. DBSCAN Clustering Algorthm:*

DBSCAN (Density-Based Spatial Clustering of Applications with Noise) [17] is a density-based data clustering algorithm because it finds a number of clusters starting from the estimated density distribution of corresponding nodes.

Given a set of $m$ data points $X = \{x_i \mid i = 1 \cdots m\}$, where each data point is a $n$-dimensional vector. The Euclidean distance between the two data points $x_p \in X$ and $x_q \in X$ is given by

$$d^2(x_p, x_q) = \left\| \sum_{k=1}^{n} x_k^p - x_k^q \right\|^2 \quad (11)$$

where, $n$ is the number of dimensions of each data point
$x_k^p$ is the value of $k^{th}$ dimensions of $x_p$
$m_k^q$ is the value of $k^{th}$ dimensions of $x_q$

In this algorithm concept of a cluster is based on the notion of "ε-neighborhood" and "density reachability". Let the ε-neighborhood of a data point $x_p$, denoted as $N_\varepsilon(x_p)$ is defined as below:

$$N_\varepsilon(x_p) = \left\{ x_q \in X \mid d^2(x_p, x_q) \leq \varepsilon \right\} \quad (12)$$

Let $\eta$ be the minimum number of points required to form a cluster. A point $x_q$ is directly density-reachable from a point $x_p$, if $x_q$ is part of ε-neighborhood of $x_p$ and if the number of points in the ε-neighborhood of $x_p$ are greater than or equal to $\eta$ as specified in (13)

$$x_q \in N_\varepsilon(x_p) \quad (13)$$
$$\left| N_\varepsilon(x_p) \right| \geq \eta$$

$x_q$ is called density-reachable from $x_p$ if there is a sequence $x_1$, … , $x_n$ of points with $x_1 = x_p$ and $x_n = x_q$ where each $x_{i+1}$ is directly density-reachable from $x_i$. Two points $x_p$ and $x_q$ are said to be density-connected if there is a point $x_o$ such that $x_o$ and $x_p$ as well as $x_o$ and $x_q$ are density-reachable.

A cluster of data points satisfies two properties: i) All the data points within the cluster are mutually density-connected. ii) If a data point is density-connected to any data point of the cluster, it is part of the cluster as well.

Input to DBSCAN algorithm are i) ε (epsilon) and ii) $\eta$, the minimum number of points required to form a cluster. The algorithm starts by randomly selecting a starting data point that has not been visited. If the ε-neighborhood of this data point contains sufficiently many points, a cluster is started. Otherwise, the data point is labeled as noise. Later this point might be found in a sufficiently sized ε-neighborhood of a different data point and hence could become part of a cluster. If a data point is found to be part of a cluster, all the data points in its ε-neighborhood are also part of that cluster and hence added to the cluster. This process continues until the cluster is completely found. Then, a new unvisited point is selected and processed, leading to the discovery of a next cluster or noise. Fig. 4 describes the DBSCAN algorithm.

```
Algorithm: DBSCAN
Input:   i) Set of m data points X={x_1, …, x_m},
         ii) ε (epsilon), the neighborhood distance and
         iii) η , the minimum number of data points
              required to form a cluster.
Output:  Set of clusters C = {c_1, …, c_k},
Steps:
1)   C = Ø.; i = 0;
2)   for each  x_p ∈ X and  x_p.visited = false
3)   begin
4)       x_p.visited = true
5)       N_p = N_ε(x_p) using (13)
6)       if |N_ε(x_p)| < η then
7)           x_p.noice = true
8)       else
9)           i = i + 1
10)          C = C ∪ c_i
11)          c_i = c_i ∪ x_p
12)          for each  x_q ∈ N
13)          begin
14)              if x_q.visited = false then
15)                  x_q.visited = true
16)                  N_q = N_ε(x_q)
17)                  if |N_ε(x_q)| < η then
18)                      N_p = N_p ∪ N_q
19)                      if x_q ∉ c_j  ∀j = 1 ≤ j ≤ i then
20)                          c_i = c_i ∪ x_q
21)                      endif
22)                  endif
23)              endif
24)          end
25)      endif
26)  end
```

Figure 5. DBSCAN Algorithm

### III. CLUSTER VALIDITY INDICES

Cluster formation using various clustering algorithms result in different clusters. Therefore it is important to perform an evaluation of the results to assess their quality. Various quality measures to evaluate the quality of the discovered clusters are described below:

#### A. Dunn's Validity Index:

Dunn's Validity Index [18] attempts to identify those cluster sets that are compact and well separated. For any number of clusters, where $c_i$ represent the *i*-cluster of such partition, the Dunn's validation index, $D$, can be calculated with the following formula:

$$D = \min_{1 \leq i \leq k} \left( \min_{i+1 \leq j \leq k} \left( \frac{dist(c_i, c_j)}{\max_{1 \leq l \leq k} diam(c_l)} \right) \right) \quad (14)$$

where

$dist(c_i, c_j)$ is distance between clusters $c_i$ and $c_j$ where

$$dist(c_i, c_j) = \min_{x_i \in c_i, x_j \in c_j} d(x_i, x_j),$$

$d(x_i, x_j)$ is distance between data points $x_i \in c_i$ and $x_j \in c_j$,

$diam(c_l)$ is diameter of cluster $c_l$ where

$$diam(c_l) = \max_{x_{l_1}, x_{l_2} \in c_l} d(x_{l_1}, x_{l_2})$$

An optimal value of the $k$ is one that maximizes the Dunn's index.

#### B. Davies-Bouldin Validity Index:

This index attempts to minimize the average distance between each cluster and the one most similar to it [19]. It is defined as:

$$DB = \frac{1}{k} \sum_{i=1}^{k} \max_{1 \leq j \leq k, j \neq i} \left( \frac{diam(c_i) + diam(c_j)}{dis(c_i, c_j)} \right) \quad (15)$$

An optimal value of the k is the one that minimizes this index.

#### C. C Validity Index:

C Index [28] is defined as:

$$C = \frac{S - S_{min}}{S_{max} - S_{min}}, \quad (16)$$

Here $S$ is the sum of distances over all pairs of objects form the same cluster. Let $m$ be the number of those pairs and $S_{min}$ is the sum of the $m$ smallest distances if all pairs of objects are considered. Similarly $S_{max}$ is the sum of the $m$ largest distances out of all pairs. The interval of the C-index values is [0, 1] and this value should be minimized.

#### D. Silhouette Validity Index

This technique computes the silhouette width for each data point, average silhouette width for each cluster and overall average silhouette width for the total data set [20]. To compute the silhouettes width of $i^{th}$ data point, following formula is used:

$$S_i = \frac{b_i - a_i}{\max(a_i, b_i)}, \quad (17),$$

where $a_i$ is average dissimilarity of $i^{th}$ data point to all other points in the same cluster; $b_i$ is minimum of average dissimilarity *of* $i^{th}$ data point to all data points in other cluster. Equation (16) results in $-1 \leq S_i \leq 1$. A value of $S_i$ close to 1 indicates that the data point is assigned to a very appropriate cluster. If $S_i$ is close to zero, it means that that data pint could be assign to another closest cluster as well because it is equidistant from both the clusters. If $S_i$ is close to –1, it means that data is misclassified and lies somewhere in between the clusters. The overall average silhouette width for the entire data set is the average $S_i$ for all data points in the whole dataset. The largest overall average silhouette indicates the best clustering. Therefore, the number of cluster with maximum overall average silhouette width is taken as the optimal number of the clusters.

*E. Rand Index*

This index [21] measures the number of pair wise agreements between the set of discovered clusters $K$ and a set of class labels $C$, is given by:

$$R = \frac{a+d}{a+b+c+d}, \qquad (18)$$

where $a$ denotes the number of pairs of data points with the same label in $C$ and assigned to the same cluster in $K$, $b$ denotes the number of pairs with the same label, but in different clusters, $c$ denotes the number of pairs in the same cluster, but with different class labels and $d$ denotes the number of pairs with a different label in $C$ that were assigned to a different cluster in $K$. The index results in $0 \leq R \leq 1$, where a value of 1 indicates that $C$ and $K$ are identical. A high value for this index generally indicates a high level of agreement between a clustering and the natural classes.

*F. Jaccard Index*

Jaccard index [22], is used to assess the similarity between different partitions of the same dataset, the level of agreement between a set of class labels $C$ and a clustering result $K$ is determined by the number of pairs of points assigned to the same cluster in both partitions:

$$J = \frac{a}{a+b+c}, \qquad (19)$$

where $a$ denotes the count of pairs of points with the same label in $C$ and assigned to the same cluster in $K$, b denotes the count of pairs with the same label, but in different clusters and c denotes the number of pairs in the same cluster, but with different class labels. The index results in $0 \leq J \leq 1$, where a value of 1 indicates that $C$ and $K$ are identical.

*G. Fowlkes-Mallows Index*

Let $K$ be the set of discovered clusters and $C$ be the set of class labels. Let $A$ be the set of all the data point pairs corresponding to the same class in $C$, and $B$ be the set of all the data point pairs corresponding to the same cluster in $K$. Then the probability that a pair of vertices which are in the same class under $C$, are also in the same cluster under $K$ is given by:

$$P(C,K) = \frac{|A| \cup |B|}{|A|}, \qquad (20)$$

It is clear that (18) is asymmetric, i.e. $P(C,K) \neq P(K,C)$, Fowlkes-Mallows Index is defined as the geometric mean of $P(C,K)$ and $P(K,C)$ [23]:

$$F(C,K) = \sqrt{P(C,K) \times P(K,C)}, \qquad (21)$$

The value of the Fowlkes-Mallows Index is between 0 and 1, and a high value means better accuracy.

IV. EXPERIMENTAL RESULTS

In order to discover the clusters that exist in user accesses sessions of a web site, we carried out a number of experiments using various clustering techniques. The Web access logs were taken from the P.A. College of Engineering, Mangalore web site, at URL http://www.pace.edu.in. The site hosts a variety of information, including departments, faculty members, research areas, and course information. The Web access logs covered a period of one month, from February 1, 2011 to February 8, 2011. Web access logs are first preprocessed to discover the user accessions. Table I depicts the results of cleaning, user identification and user session identification. Details of our preprocessing strategies can be found in [24].

TABLE I:

RESULTS OF CLEANING AND USER IDENTIFICATION

| Items | Count |
|---|---|
| Initial No of Log Entries | 12744 |
| Log Entries after Cleaning | 11995 |
| No. of site ULRs accessed | 116 |
| No of Users Identified | 16 |
| No. of User Sessions Identified | 206 |

The preprocessed user sessions are clustered using five different clustering algorithms. The details of the experiments are given below:

1. Conducted the multiple runs of Leader algorithm by selecting the input parameter ε (Dissimilarity Threshold) ranging from ε = 0.5,…, 3.5 in steps of 0.5.

2. Conducted multiple runs of k-Means algorithm by selecting the input parameter $k$ (number of clusters) ranging from k = 2, …, 25.

3. Conducted multiple runs of k-Means algorithm by selecting the input parameter $k$ (number of clusters) ranging from k = 2, …, 25.

4. Conducted multiple runs of Single Link Hierarchical Agglomerative clustering algorithm by selecting the termination condition for the number of clusters $k$ ranging from k = 2, …, 25.

5. Conducted multiple runs of DBSCAN algorithm by selecting the input parameter ε (neighborhood distance) ranging from ε = 0.5, …, 3.5 in steps of 0.5. The other parameter η, which indicates the minimum no. of points in a cluster is set in a range from η = 2, …, 10.

In order to evaluate the accuracy of the cluster formation, we computed the clustering error SSE (sum of the squared error), for each of the above runs. Moreover, since the above clustering algorithms result in different cluster formation, it is important to assess their quality. We evaluated our results based on the following validity indices: i) Dunn's Index, ii) Davies Bouldin, iii) Jaccard Index, iv) C Index, v) Rand Index, vi) Silhouette Index and vii) Fowlkes-Mallows Index. Finally, to evaluate the time performance, execution times are recorded in milliseconds. Table II describes the various performance measures and validity indices corresponding to each of the above mentioned clustering algorithms. They are further explained in the following subsections:

*1) Evaluation of Cluster Validity Using Dunn's Index:*

Fig. 7 shows the graph plot of Dunn's validity index values as a function of the number of clusters $k$. The main goal of the measure is to maximize the inter-cluster distances and minimize the intra-cluster distances. From the graph it is clear that we are getting best performance using Hierarchical clustering in terms of the Dunn's validity index.

*2) Evaluation Using Davies Bouldin Index:*

Fig. 8 shows the graph plot of Davies Bouldin (DB) validity index values as a function of the number of clusters $k$. This index attempts to minimize the average distance between each cluster and the one most similar to it. From the graph it is clear that Single link clustering provides the minimum values for this index, thus outperforming the other techniques.

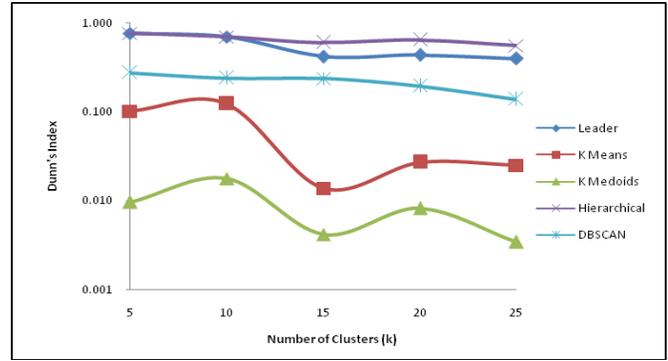

Figure 6. Dunn's Index Vs. No. of Clusters $k$

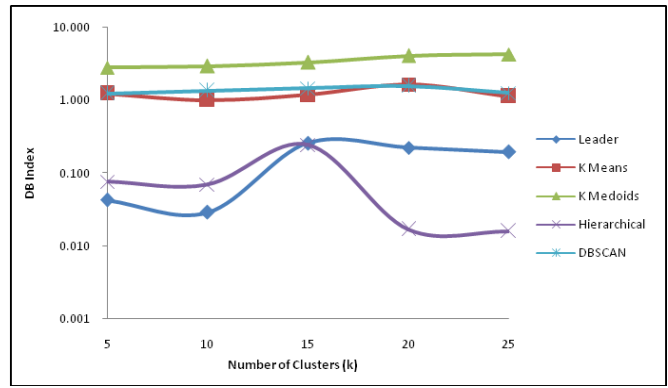

Figure 7. Davies Bouldin Index Vs. No. of Clusters $k$

TABLE II  EVALUATION OF PERFORMANCE AND VALIDITY FOR VARIOUS CLUSTERING TECHNIQUES

| Technique | No. of Clusters | Dunn's Index | DB Index | Jaccard Index | C Index | Rand Index | Fowlkes Mallows Index | Silhouette Index | SSE | Execution Time |
|---|---|---|---|---|---|---|---|---|---|---|
| Leader | 05 | 0.7741 | 0.0431 | 0.9614 | 0.0023 | 0.9614 | 0.9805 | 0.75434 | 128.96 | 1 |
|  | 10 | 0.7039 | 0.0293 | 0.9143 | 0.0091 | 0.9143 | 0.9562 | 0.59389 | 104.43 | 2 |
|  | 15 | 0.4216 | 0.2587 | 0.8594 | 0.0153 | 0.8594 | 0.9271 | 0.60250 | 88.18 | 1 |
|  | 20 | 0.4391 | 0.2253 | 0.8149 | 0.0159 | 0.8149 | 0.9027 | 0.60417 | 76.11 | 2 |
|  | 25 | 0.4011 | 0.1950 | 0.7546 | 0.0159 | 0.7546 | 0.8687 | 0.54886 | 61.44 | 2 |
| k-Means | 05 | 0.1018 | 1.2451 | 0.8607 | 0.1085 | 0.8607 | 0.9277 | 0.4640 | 151.56 | 45 |
|  | 10 | 0.1236 | 1.0075 | 0.6945 | 0.1297 | 0.6945 | 0.8334 | 0.3392 | 117.61 | 66 |
|  | 15 | 0.0137 | 1.1885 | 0.5240 | 0.1171 | 0.5240 | 0.7239 | 0.5531 | 105.83 | 169 |
|  | 20 | 0.0270 | 1.6541 | 0.3995 | 0.1020 | 0.3995 | 0.6321 | 0.5287 | 109.64 | 141 |
|  | 25 | 0.0250 | 1.1352 | 0.4594 | 0.0875 | 0.4594 | 0.6778 | 0.4883 | 78.82 | 242 |
| k-Medoids | 05 | 0.0097 | 2.8323 | 0.6162 | 0.3855 | 0.6162 | 0.7850 | 0.13195 | 162.95 | 7 |
|  | 10 | 0.0178 | 2.9726 | 0.6557 | 0.2444 | 0.6557 | 0.8097 | 0.33236 | 137.20 | 3 |
|  | 15 | 0.0042 | 3.3482 | 0.4373 | 0.2685 | 0.4373 | 0.6613 | 0.26599 | 117.03 | 6 |
|  | 20 | 0.0082 | 4.0912 | 0.2747 | 0.2072 | 0.2747 | 0.5241 | 0.05941 | 118.30 | 6 |
|  | 25 | 0.0034 | 4.2914 | 0.1846 | 0.2508 | 0.1846 | 0.4297 | 0.15755 | 118.14 | 5 |
| Single Link | 05 | 0.7741 | 0.0771 | 0.9614 | 0.0023 | 0.9614 | 0.9805 | 0.73474 | 128.96 | 10383 |
|  | 10 | 0.7039 | 0.0705 | 0.9143 | 0.0091 | 0.9143 | 0.9562 | 0.61199 | 104.43 | 10202 |
|  | 15 | 0.6074 | 0.2459 | 0.8594 | 0.0182 | 0.8594 | 0.9270 | 0.64592 | 87.61 | 10690 |
|  | 20 | 0.6519 | 0.0169 | 0.8236 | 0.0115 | 0.8236 | 0.9075 | 0.54089 | 75.44 | 10189 |
|  | 25 | 0.5613 | 0.0160 | 0.7801 | 0.0146 | 0.7801 | 0.8832 | 0.54421 | 65.52 | 10308 |
| DBSCAN | 05 | 0.2774 | 1.2517 | 0.6547 | 0.1285 | 0.6547 | 0.8091 | 0.20504 | 766.90 | 41 |
|  | 10 | 0.2402 | 1.3666 | 0.4829 | 0.1985 | 0.4829 | 0.6949 | 0.13979 | 675.44 | 12 |
|  | 15 | 0.2387 | 1.4780 | 0.4778 | 0.2258 | 0.4778 | 0.6912 | 0.15142 | 857.78 | 13 |
|  | 20 | 0.1961 | 1.5917 | 0.2067 | 0.4682 | 0.2067 | 0.4547 | 0.30353 | 805.88 | 21 |
|  | 25 | 0.1387 | 1.2594 | 0.4130 | 0.6606 | 0.4130 | 0.6427 | 0.30596 | 861.24 | 22 |

*3) Evaluation Using C Index:*

Fig. 9 shows the graph plot of C validity index values as a function of the number of clusters *k*. C-index values should be minimized. Single link and Leader clustering provide the minimum values for this index and hence outperform the other techniques.

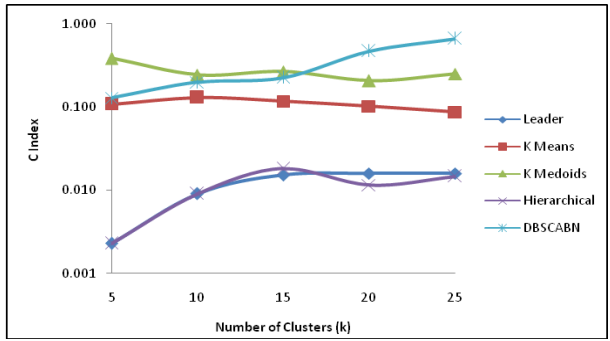

Figure 8.   C Validity Index Vs. No. of Clusters *k*

*4) Evaluation Using Jaccard Index:*

Fig. 10 shows the graph plot of Jaccard validity index values as a function of the number of clusters *k*. The Jaccard index values range between 0 and 1, and higher values indicates the better cluster validity. From the graph it is clear that Single link agglomerative hierarchical clustering and Leader clustering (with dissimilarity threshold $\varepsilon$ = 1.0) provide the maximum values for this index and hence outperform the other techniques.

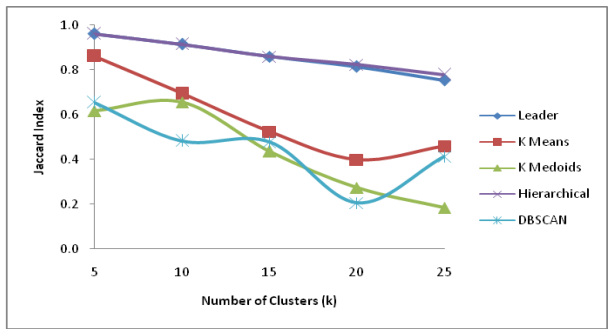

Figure 9.   Jaccard Validity Index Vs. No. of Clusters *k*

*5) Evaluation Using Silhouette Index:*

Fig. 11 shows the graph plot of Silhouette validity index values as a function of the number of clusters *k*. Silhouetter values range between -1 and 1. A value close to 1 indicates that the data point is assigned to a very appropriate cluster. A value is close to zero means that that data pint could be assign to another closest cluster as well because it is equidistant from both the clusters. If the value is close to –1, it means that data is misclassified and lies somewhere in between the clusters. Our results show that all the techniques generate the values between 0 and 1. Single link agglomerative hierarchical clustering and Leader clustering (with dissimilarity threshold $\varepsilon$ = 1.0) provide values closer to 1 and hence outperform the other techniques.

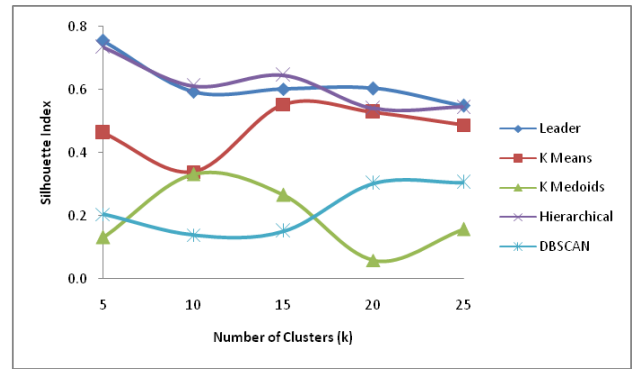

Figure 10. Silhouette Validity Index Vs. No. of Clusters *k*

*6) Evaluation Using Fowlkes-Mallows Index:*

Fig. 11 shows the graph plot of Fowlkes Mallows index as a function of the number of clusters *k*. A high value of this index means better accuracy. Our results show that Single link and Leader clustering provide the maximum values for this index and hence outperform the other techniques.

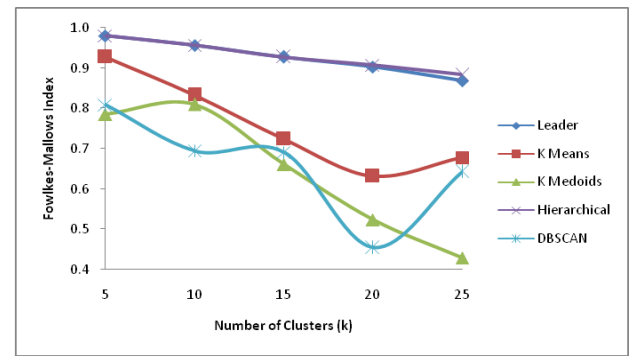

Figure 11. Fowlkes-Mallows Validity Index Vs. No. of Clusters *k*

*7) Evaluation Using Rand Index:*

Fig. 12 shows the graph plot of Rand validity index values as a function of the number of clusters *k*. The index values range between 0 and 1. A high value for this index indicates a high level of agreement between a clustering and the natural classes. Our results show that Single link and Leader clustering provide the maximum values for this index and hence outperform the other techniques.

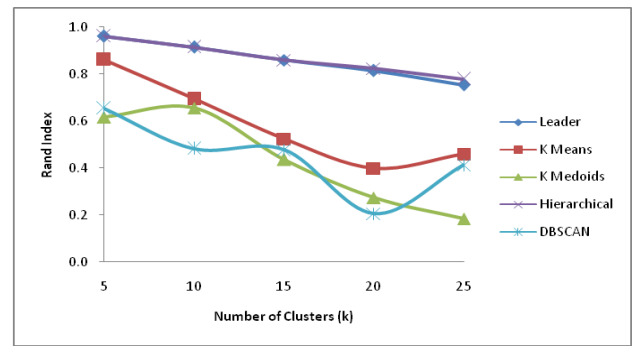

Figure 12. Rand Index Vs. No. of Clusters *k*

*8) Evaluation of the Clustering Error (SSE):*

Graph plot in Fig. 6, displays the clustering error (SSE) for various clustering algorithms as a function of the number of clusters, *k*.

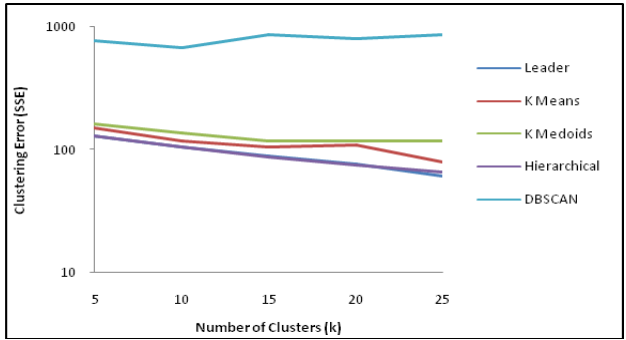

Figure 13. Clustering Error (*J*) Vs. No. of Clusters *k*

Graph shows that accuracy of the k-Means algorithm is better than the k-Medoids algorithms based on clustering error. Single link agglomerative hierarchical clustering and Leader clustering (with dissimilarity threshold ε = 1.0) provide the best accuracy. The performance of DBSCAN algorithm is very poor in terms of SSE, because DBSCAN results in arbitrary shape of clusters which may not be globular and SSE is more suitable measure for those algorithms which result in globular clusters.

*9) Evaluation of Time Performance:*

Fig. 14 shows the graph plot of Execution Time in millikseconds as a function of the number of clusters *k*. Our results show that Single link agglomerative hierarchical clustering algorithm is the slowest and Leader clustering (with dissimilarity threshold ε = 1.0) is the fastest.

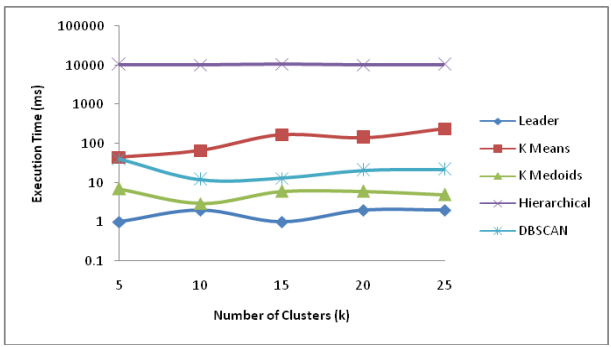

Figure 14. Execution Time (ms) Vs. No. of Clusters *k*

## V. CONCLUSION

In this paper we have we provided a detailed overview of various techniques used for clustering the users' navigational sessions. We described the underlying mathematical model and algorithm details related to the implementation of these clustering algorithms. In order to evaluate the validity of the clusters formed, we discussed various validity indices. We also discussed sum of the squared error as a measure of accuracy of the clustering. Time performance is evaluated by recording the execution timings of various algorithms. From the results presented in the previous section, we conclude the following points related with user session clustering.

- Single link agglomerative hierarchical clustering provides the best accuracy in terms of sum of the squared error (SSE). Accuracy of the k-Means algorithm is better than that of the k-Medoids algorithm. The performance of DBSCAN algorithm is very poor in terms of SSE, because DBSCAN results in arbitrary shape of clusters which may be non-globular and SSE is a more suitable measure for globular clusters.

- Single link agglomerative hierarchical clustering gives best results for all the cluster validity measures described above followed by the Leader clustering (with dissimilarity threshold ε = 1.0). k-Means algorithm outperforms the k-Medoids. DBSCAN gives better performance than k-Means with respect to Dunn's and DB indices, but for all other indices k-Means gives better result than DBSCAN.

- Our result shows that Leader clustering algorithm gives the best time performance. DBSCAN and k-medoids algorithms provide reasonably better time performance than that k-Means algorithm. The reason for this is, for an optimal solution k-Means algorithm performs multiple runs and computes the distance between every data object and its corresponding cluster center thus taking more execution Time. As far as time measure is concerned, Single Link Hierarchical clustering algorithm takes the execution time much more than all other algorithms and thus provides the worst time performance.

AUTHORS PROFILE

**Zahid Ansari** is a Ph.D. candidate in the Department of CSE, Jawaharlal Nehru Technical University, India. He received his ME from Birla Institute of Technology, Pilani, India. He has worked at Tata Consultancy Services (TCS) where he was involved in the development of cutting edge tools in the field of model driven software development. His areas of research include data mining, soft computing and model driven software development. He is currently with the P.A. College of Engineering, Mangalore as a Faculty. He is also a member of ACM.

**A.Vinaya Babu** is working as Director of Admissions and Professor of CSE at J.N.T. University Hyderabad, India. He received his M.Tech. and PhD in Computer Science Engineering from JNT University, Hyderabad. He is a life member of CSI, ISTE and member of FIE, IEEE, and IETE. He has published more than 35 research papers in International/National journals and Conferences. His current research interests are algorithms, information retrieval and data mining, distributed and parallel computing, Network security, image processing etc.

**Mohammad Fazle Azeem** is working as Professor and Director of department of Electronics and Communication Engineering, P.A. College of Engineering, Mangalore. He received his B.E. in electrical engineering from M.M.M. Engineering College, Gorakhpur, India, M.S. from Aligarh Muslim University, Aligarh, India and Ph.D. from Indian Institute of Technology (IIT) Delhi, India. His interests include robotics, soft computing, clustering techniques, application of neuro-fuzzy approaches for the modeling, and control of dynamic system such as biological and chemical processes.

**Waseem Ahmed** is a Professor in CSE at P.A. College of Engineering, Mangalore. He obtained his BE from RVCE, Bangalore, MS from the University of Houston, USA and PhD from the Curtin University of Technology, Western Australia. His current research interests include multicore/multiprocessor development for HPC and embedded systems, and data mining. He has been exposed to academic/work environments in the USA, UAE, Malaysia, Australia and India where he has worked for more than a decade. He is a member of the IEEE.